# On Supervised Selection of Bayesian Networks

**Petri Kontkanen    Petri Myllymäki**
**Tomi Silander    Henry Tirri**
Complex Systems Computation Group (CoSCo)
P.O.Box 26, Department of Computer Science
FIN-00014 University of Helsinki, Finland
http://www.cs.Helsinki.FI/research/cosco/

## Abstract

Given a set of possible models (e.g., Bayesian network structures) and a data sample, in the unsupervised model selection problem the task is to choose the most accurate model with respect to the domain joint probability distribution. In contrast to this, in *supervised model selection* it is a priori known that the chosen model will be used in the future for prediction tasks involving more "focused" predictive distributions. Although focused predictive distributions can be produced from the joint probability distribution by marginalization, in practice the best model in the unsupervised sense does not necessarily perform well in supervised domains. In particular, the standard marginal likelihood score is a criterion for the unsupervised task, and, although frequently used for supervised model selection also, does not perform well in such tasks. In this paper we study the performance of the marginal likelihood score empirically in supervised Bayesian network selection tasks by using a large number of publicly available classification data sets, and compare the results to those obtained by alternative model selection criteria, including empirical crossvalidation methods, an approximation of a supervised marginal likelihood measure, and a supervised version of Dawid's *prequential* (predictive sequential) principle. The results demonstrate that the marginal likelihood score does not perform well for supervised model selection, while the best results are obtained by using Dawid's prequential approach.

## 1 INTRODUCTION

One of the recent active areas in Bayesian network research has focused on the problem of learning the Bayesian network structure, given a sample of data from the problem domain. In most cases the approach for solving this problem is based on the idea of separating the model search and the scoring of the different models. In this paper we focus on the scoring aspect of the model selection problem; in other words, we are interested in scoring functions, or *(model selection) criteria* in terms of (Heckerman & Meek, 1997), which define what networks are considered "good" models.

The most commonly used model selection criterion in the Bayesian network domain is the (unsupervised) marginal likelihood, sometimes also called the evidence measure. By making certain technical assumptions, this criterion can be computed efficiently, as described in (Cooper & Herskovits, 1992; Heckerman, Geiger, & Chickering, 1995). Although this score can be shown to possess some desirable theoretical properties (see (Bernardo & Smith, 1994; Merhav & Feder, 1998)), the results hold only in specific situations. Regardless of this, marginal likelihood is typically used also in model selection tasks where the optimality results no longer hold. One frequently occurring situation is the supervised model selection task, where the goal is to choose a model for a particular prediction task, for example, in a classification problem. Obviously in this case predictive distributions can be produced from the joint probability distribution corresponding to the model selected by the marginal likelihood score. However, as discussed e.g., in (Heckerman & Meek, 1997), the marginal likelihood score is basically an unsupervised learning criterion, and as such not necessarily optimal for supervised model selection tasks. In this paper we study this claim in several classification domains, and demonstrate empirical results to support it.

Another problem in using the marginal likelihood score for model selection is the fact that the score is inherently optimized for a particular loss function, i.e., for logarithmic loss. The predictive distribution determined by a selected Bayesian network structure can, by using decision theory, be used for minimizing any given loss function. However, it should be observed that if the loss function is known in advance before choosing the model, for optimal performance it should already be taken into account in the model selection decision (see the discussion in (Dawid, 1992)). Modifying the marginal likelihood approach for arbitrary loss functions is by no means straightforward (Grünwald, 1998).

On the other hand, focused prediction tasks, such as classification, can be seen to define a focused loss function, so in general we can say that unsupervised learning deals with the logarithmic loss in connection with the joint probability distribution, while supervised learning is related to other types of loss functions. However, finding a more formal definition for the difference of unsupervised and supervised learning seems to be a difficult task. Instead of making an attempt towards this, in this paper we focus on the most typical supervised model selection problem, the task of selecting a model in a classification domain where the goal is to produce a predictive distribution for a single discrete class variable. In our experiments reported in Section 3.2, the predictive accuracy of the selected classifier model is measured by using both the 0/1-loss and the logarithmic loss on this classification predictive distribution. Our goal was to empirically demonstrate that although frequently used in practice also in this type of classification domains, the unsupervised marginal likelihood model selection score does not perform well in supervised model selection tasks.

As an alternative to the unsupervised marginal likelihood model selection score, we consider several other model selection criteria, including Dawid's prequential approach (Dawid, 1984), empirical crossvalidation methods (Stone, 1974; Geisser, 1975), and the supervised marginal likelihood approximation discussed in (Kontkanen, Myllymäki, Silander, & Tirri, 1998). As opposed to the unsupervised marginal likelihood, these criteria can be easily modified for different loss functions. Dawid's prequential approach can also be shown to possess certain elegant asymptotic properties, but for some reason this method has been rarely used in practice. For our purposes, we modify the prequential score for classification domains by using Cox's partial marginal likelihood principle (Cox, 1975) as suggested in (Dawid, 1991). In (Spiegelhalter, Dawid, Lauritzen, & Cowell, 1993), the resulting criterion was called the *conditional node monitor*. The third cri-terion, the supervised modification of the marginal likelihood score, was called the *class sequential criterion* in (Heckerman & Meek, 1997). The various scoring methods used are described in more detail in Section 2.2.

The criteria discussed above were empirically evaluated in different supervised model selection domains by using 18 public domain real-world classification data sets. As mentioned earlier, in this paper we concentrate on the problem of determining a good criterion. Consequently, in order to eliminate the effect of model search, we wanted to restrict the model family in such a way that in the empirical tests it would be possible to exhaustively score all the models in the chosen family. For this reason, the model family used in the tests reported in Section 3.2 was chosen to be the set containing all Bayesian network structures with the property that all the arcs start from the root node representing the class variable.

The reasons for choosing this subset of Bayesian networks for this set of experiments are twofold. First of all, although the number of possible models in this setup is equal to the number of possible subsets of the domain variables, and hence still quite large, by restricting ourselves to datasets with less than 15 variables we were able to perform the exhaustive model search by using a network of Linux workstations. Secondly, all the models in the selected model family can be seen as "pruned Naive Bayes" models, which means that the model selection problem in this case be regarded as a feature selection problem in the Naive Bayes classifier domain. As the Naive Bayes model is one of the most well-known models in classification domains, and the empirical results show that some of the selected model selection criteria consistently improve the predictive accuracy of the model, the results are of practical relevance.

We would like to emphasize that all the model selection criteria used in this paper are directly applicable to Bayesian networks in general, not just the Naive Bayes model or its variants. Therefore the experiments could be easily repeated with general Bayesian network structures, but in this case exhaustive model search would no longer be possible because of the huge number of possible models, and some kind of a search algorithm would be needed for choosing candidate models. Consequently, in this case it would be difficult to say whether the differences in empirical results would be caused the properties of different model selection criteria, or by the bias caused by the search algorithm used.

When compared to a related empirical study reported in (Friedman, Geiger, & Goldszmidt, 1997), the work presented here differs in several aspects. First, the emphasis on this paper is strictly on the problem of selecting the Bayesian network structure, while (Friedman et al., 1997) consider the problem of finding both the model structure and the model parameters. However, this difference is not crucial in the Bayesian network modeling framework where setting the parameter values to their expected values produces a predictive distribution which is equal to that obtained by integrating over the parameters (Heckerman et al., 1995). Second, although both studies deal with extensions of the Naive Bayes model, (Friedman et al., 1997) try to relax the unrealistic Naive Bayes assumptions by adding arcs between the leaf nodes of the network (the TAN model), while in this paper we prune the Naive Bayes network by removing arcs between the root node and the leaves. Third, although (Friedman et al., 1997) recognize the need for supervised model selection criteria, for computational reasons they resort to unsupervised methods and focus on comparing results between different model families, while the goal of this paper is to compare the performance of supervised and unsupervised model selection criteria in supervised domains by using a single (fixed) set of models.

Our empirical results demonstrate that unsupervised marginal likelihood really shows mediocre performance in supervised model selection tasks. In addition, the supervised marginal likelihood modification also performs poorly, but we argue that this is caused by the crude approximation method used, not by the properties of the criterion itself (see the discussion in (Kontkanen et al., 1998)). Empirical crossvalidation methods perform relatively well, which is not surprising as they can be viewed as approximations of the supervised marginal likelihood, as demonstrated in Section 2.2.3. Most interestingly, the best results were obtained by using a supervised version of Dawid's prequential approach. However, one caveat is that the performance of the prequential method depends on the ordering of the data sequence, hence a more thorough study on this aspect is an obvious topic for future work. The empirical supervised model selection test setup used in the tests is described in Section 3.1, and the results are summarized in Section 3.2.

## 2 SUPERVISED MODEL SELECTION

### 2.1 THE PROBLEM

Let $\mathcal{D} = \mathbf{x}^N$ denote a matrix of $N$ vectors each consisting values of $n$ random variables $X_1, \ldots, X_n$. For simplicity, in the sequel we will assume the random variables $X_i$ to be discrete. By a model $M$ we mean here a parametric model form so that each parameterized instance $(M, \theta)$ of the model produces a probability distribution $P(X_1, \ldots, X_n | M, \theta)$ on the space of possible data vectors $\mathbf{x}$. To make our presentation more concrete, for the remainder of the paper we assume the models $M$ to represent different *Bayesian network structures* (for an introduction to Bayesian network models, see e.g., (Pearl, 1988)).

Given a selection $\mathcal{F} = \{M_1, \ldots, M_m\}$ of possible models (Bayesian network structures), and a data sample $\mathcal{D}$, in the (unsupervised) model selection problem, the task is to choose a model $M$ so that the resulting predictive distribution

$$P(X_1, \ldots, X_n | \mathcal{D}, M)$$
$$= \int P(X_1, \ldots, X_n | \mathcal{D}, M, \theta) P(\theta | \mathcal{D}, M) d\theta$$

yields the most accurate predictions.

Two observations are now in order. First, the best model in the above sense depends on how we measure the accuracy of the resulting predictive distribution; in other words, on the loss function to be used for this purpose. Second, although it is intuitively appealing to think of the data $\mathcal{D}$ as a random sample from some "true" but unknown probability distribution, the model selection problem can be addressed without such an assumption. An excellent survey on the theoretical aspects of this topic can be found in (Merhav & Feder, 1998).

By *supervised model selection* we mean a situation where the domain variables can be partitioned into two separate sets $\mathcal{U} = \{U_1, \ldots, U_n\}$ and $\mathcal{V} = \{V_1, \ldots, V_{n'}\}$, and we know a priori that *all* future prediction tasks involve predicting the values of variables in $\mathcal{V}$, given the values of variables in set $\mathcal{U}$. For notational simplicity, let us in the sequel assume that the set $\mathcal{V}$ consists of a single variable $V$, in which case we are dealing with classification problems.

In the supervised classification framework described above, the goal in the model selection is thus to choose the model $M$ which yields the most accurate classifications with respect to the loss function used. It is now important to see that although the joint probability distribution $P(\mathbf{x}|M)$ can be used for producing the required classification probability distribution $P(\mathbf{v}|\mathbf{u}, M)$,

$$P(\mathbf{v}|\mathbf{u}, M) \propto P(\mathbf{v}, \mathbf{u}|M) = P(\mathbf{x}|M),$$

the model $M$ producing the most accurate predictive distribution in the joint probability estimation sense does not necessarily result in the most accurate classification probability distribution.

## 2.2  THE METHODS

### 2.2.1  Unsupervised and Supervised Marginal Likelihood

One way to look at the model selection problem is to regard $\mathcal{F}$ as a random variable (with possible values $M_1, \ldots, M_m$), and choose the model maximizing the posterior probability $P(M|\mathcal{D})$. Assuming all the models to be equally probable a priori, this leads to choosing the model $M^*$ maximizing the *marginal likelihood* or the *evidence* of the data $\mathcal{D}$:

$$M^* = \arg\max_M P(M|\mathcal{D}) = \arg\max_M P(\mathcal{D}|M)$$

$$(1) \qquad = \arg\max_M \int P(\mathcal{D}|M, \theta) P(\theta|M) d\theta.$$

We see that the marginal likelihood measure depends on the prior distribution $P(\theta|M)$ defined on the model parameters. This prior can either be regarded as a formalization of our prior domain knowledge, which leads to interesting questions about the compatibility of different priors (Heckerman & Geiger, 1995; Cowell, 1992), or as a technical parameter representing no such information. In the latter case, it can be shown that a certain prior known as Jeffreys' prior (Jeffreys, 1946; Berger, 1985) can be given strong theoretical justification from the predictive performance point of view with respect to the so called minimax loss formulation (Rissanen, 1996; Grünwald, 1998). Some empirical results concerning the effect of Jeffreys' prior on predictive accuracy can be found in (Grünwald, Kontkanen, Myllymäki, Silander, & Tirri, 1998). In the remainder of this paper we do not address the problem of choosing the prior distributions, but use uniform priors for the parameters.

The marginal likelihood measure (1) is the most commonly used model selection method in Bayesian network learning. Nevertheless, although marginal likelihood is an *unsupervised* model selection method in the sense that the chosen model $M^*$ represents well the *joint* distribution $P(X_1, \ldots, X_n|M)$, the method is frequently used also for supervised model selection tasks where it may result in poor results. For this reason, in (Kontkanen et al., 1998) we suggested the following modification of the marginal likelihood to be used in supervised model selection tasks:

$$(2) \quad M^* = \arg\max_M P(\mathbf{v}^N \mid \mathbf{u}^N, M)$$

$$= \arg\max_M \int P(\mathbf{v}^N \mid \mathbf{u}^N, \theta, M) P(\theta|\mathbf{u}^N, M) d\theta.$$

Unfortunately, using this supervised marginal likelihood score is generally not computationally feasible even in the cases where the unsupervised marginal likelihood (1) could be computed efficiently. This can be

seen by observing that

$$P(\mathbf{v}^N \mid \mathbf{u}^N, M) = \frac{P(\mathbf{v}^N, \mathbf{u}^N \mid M)}{P(\mathbf{u}^N \mid M)}$$

$$(3) \qquad = \frac{P(\mathbf{v}^N, \mathbf{u}^N \mid M)}{\sum_{\mathbf{v}^N} P(\mathbf{v}^N, \mathbf{u}^N \mid M)}.$$

The sum in the denominator goes over all the possible configurations of vector $\mathbf{v}^N$, which are exponential in number. Approximating this sum by the Cheeseman-Stutz measure (Cheeseman & Stutz, 1996) leads to a criterion (see (Kontkanen et al., 1998)) where the supervised marginal likelihood (2) is replaced by a single supervised marginal likelihood

$$(4) \qquad P(\mathbf{v}^N \mid \mathbf{u}^N, M) \approx \prod_{i=1}^N P(\mathbf{v}^N \mid \mathbf{u}^N, \hat{\theta}, M),$$

where the parameters $\hat{\theta}$ maximize the (unsupervised) parameter posterior probability $P(\theta|\mathbf{v}^N, \mathbf{u}^N, M)$. In (Kontkanen et al., 1998) it was noted that the derivation of this approximation does not suggest using the supervised maximum likelihood parameters maximizing the supervised likelihood $P(\mathbf{v}^N \mid \mathbf{u}^N, \theta, M)$, which are generally different from the (unsupervised) maximum likelihood parameters (see the discussion in (Friedman et al., 1997)).

### 2.2.2  Prequential Approaches

In Dawid's *prequential* (predictive sequential) approach for statistical validation of models (Dawid, 1984), alternative models are compared by measuring their cumulative loss, so the validation score is computed through a sequential updating of the predictive distribution.

From the rules of probability theory it follows that

$$-\log P(\mathcal{D}|M) = -\log \prod_{i=1}^N P(\mathbf{x}_i|\mathbf{x}^{i-1})$$

$$(5) \qquad = \sum_{i=1}^N -\log P(\mathbf{x}_i|\mathbf{x}^{i-1}),$$

from where it is easy to see that maximizing the marginal likelihood leads to choosing the model minimizing the cumulative sequential logarithmic loss (with respect to sample $\mathcal{D}$). This means that marginal likelihood can be seen as a special case of the prequential approach for model selection. Furthermore, as noted in (Dawid, 1992), from the information-theoretic point of view the prequential approach can be regarded as a predictive coding system discussed in, e.g., (Rissanen, 1989). However, it should be noted that in contrast to marginal likelihood and information-theoretic approaches, which are closely linked to a specific loss function, the logarithmic loss, Dawid's prequential principle can be modified for any loss function.

The prequential model selection principle is usually described in the unsupervised model selection domain. For the supervised case, the approach has to be modified accordingly. Here we follow the suggestion given in (Dawid, 1991), and use the partial marginal likelihood measure described in (Cox, 1975). The idea is based on the observation that the marginal likelihood can be factorized into two products as follows:

$$
\begin{aligned}
P(\mathcal{D} \mid M) &= P(\mathbf{v}^N, \mathbf{u}^N \mid M) \\
&= \prod_{i=1}^{N} P(v_i, u_i \mid \mathbf{v}^{i-1}, \mathbf{u}^{i-1}, M) \\
(6) &= \prod_{i=1}^{N} P(v_i \mid \mathbf{v}^{i-1}, \mathbf{u}^i, M) \\
&\quad \cdot \prod_{i=1}^{N} P(u_i \mid \mathbf{v}^{i-1}, \mathbf{u}^{i-1}, M).
\end{aligned}
$$

Of these two products, the first one was called the *partial (marginal) likelihood* in (Cox, 1975) and *conditional node monitor* in (Spiegelhalter et al., 1993). We see that if we use the partial marginal likelihood as a basis for a prequential scoring function, this results in a sequential process where at time $i$, the classification predictive distribution

$$
P(V_i \mid \mathbf{v}^{i-1}, \mathbf{u}^i, M) = P(V_i \mid \mathbf{v}^{i-1}, \mathbf{u}^{i-1}, u_i, M)
$$

is computed by using the information preceding $v_i$ in the matrix $\mathcal{D}$ (assuming that the values of $V$ are stored in the last column of $\mathcal{D}$). Consequently, this approach produces the following principle for model selection:

$$
(7) \qquad M^* = \arg\max_M \prod_{i=1}^{N} P(\mathbf{v}_i \mid \mathbf{v}^{i-1}, \mathbf{u}^i, M).
$$

Please note that the model selected by the partial marginal likelihood approach (7) is usually not the same as the model selected by the supervised marginal likelihood approach (2), even if we could compute the supervised marginal likelihood accurately. It is also important to notice that unlike with the unsupervised case (5), the value of the partial marginal likelihood $\prod_{i=1}^{N} P(\mathbf{v}_i \mid \mathbf{v}^{i-1}, \mathbf{u}^{i-1}, M)$ depends on the ordering of the data. We return to this question in Section 3.1.

### 2.2.3 Crossvalidation

A frequently used empirical criterion for estimating the predictive accuracy of a model is provided by the *crossvalidation method* (Stone, 1974; Geisser, 1975). In this scheme, the training data is partitioned into $k$ subsets of equal size, and each subset is used in turn as a validation set while the union of the other subsets forms the data from which the predictive model is constructed. At each step, the loss associated with the resulting predictive model is computed by using the validation data, and the final estimate of the expected loss is the "crossvalidated" average over the $k$ loss values. An extreme special case of the algorithm is the *leave-one-out crossvalidation* method, where a dataset of size $N$ is partitioned into $N$ subsets, each containing a single data vector. With the logarithmic loss, the leave-one-out crossvalidation model selection criterion is thus

$$
\begin{aligned}
(8) \quad M^* = & \\
& \arg\max_M \sum_{i=1}^{N} -\log P(v_i \mid \mathbf{v}^{i-1}, v_{i+1}, \ldots, v_N, \mathbf{u}^N, M).
\end{aligned}
$$

The crossvalidation method has several appealing properties: the algorithm is easy to implement, is computationally feasible (with moderate size datasets), can be used with different loss functions, and has basically only one parameter ($k$). Nevertheless, the theoretical properties of the method are not yet fully understood. What is more, the results with $k$-fold crossvalidation (with $k < N$) seem to be highly dependent on the way the data is partitioned into the $k$ folds (Kontkanen, Myllymäki, & Tirri, 1996). Of course, with leave-one-out crossvalidation this effect does not appear, and with $k$-fold crossvalidation we can try to avoid this problem for example by repeating the algorithm over several different data partitionings and averaging over the obtained results.

In practice crossvalidation often works quite well. We can partly explain this empirical observation by noting the following connection between crossvalidation and the supervised marginal likelihood (3). First of all, it is easy to see that the supervised marginal likelihood can be factorized as follows:

$$
\begin{aligned}
-\log P(\mathbf{v}^N \mid \mathbf{u}^N, M) &= -\log \prod_{i=1}^{N} P(v_i \mid \mathbf{v}^{i-1}, \mathbf{u}^N, M) \\
(9) &= \sum_{i=1}^{N} -\log P(v_i \mid \mathbf{v}^{i-1}, \mathbf{u}^N, M),
\end{aligned}
$$

where $v_i$ denotes the value of $V$ on row $i$, i.e., the $i$th component of vector $\mathbf{v}^N$. The last term in the above sum, $-\log P(v_N \mid \mathbf{v}^{N-1}, \mathbf{u}^N, M)$, is the same term as the last term in the sum in (8). In order to compute the second to the last term in (9), $-\log P(v_{N-1} \mid \mathbf{v}^{N-2}, \mathbf{u}^N, M)$, we need to marginalize over the values of $V_N$:

(10)  $P(v_{N-1} \mid \mathbf{v}^{N-2}, \mathbf{u}^N, M)$

$= \sum_{V_N} P(v_{N-1} \mid V_N, \mathbf{v}^{N-2}, \mathbf{u}^N, M) P(V_N \mid \mathbf{v}^{N-2}, \mathbf{u}^N, M).$

This sum can be approximated by assuming that the value $v_N$ actually found in the data for the class variable $V$ on row $N$ gets probability one, while all the other values get probability zero: if $P(V_N \mid \mathbf{v}^{N-2}, \mathbf{u}^N, M) = 1$ for $V_N = v_N$, the sum (10) reduces to a single term, and we get:

(11)  $P(v_{N-1} \mid \mathbf{v}^{N-2}, \mathbf{u}^N, M)$

$\approx P(v_{N-1} \mid v_N, \mathbf{v}^{N-2}, \mathbf{u}^N, M),$

which is exactly what is computed in the second to the last step of leave-one-out crossvalidation. Continuing this line of reasoning, we see that leave-one-out-crossvalidation can be regarded as an approximation of the factorization (9), where each marginalizing sum is replaced by a single term. The $k$-fold crossvalidation scheme can be given a similar interpretation by using a different factorization.

## 3  EMPIRICAL RESULTS

### 3.1  TEST SETUP

For our supervised model selection experiments, we used publicly available classification datasets from the UCI data repository (Blake, Keogh, & Merz, 1998). As discussed earlier, we wanted to eliminate the effects of the model search procedure from our results, hence we restricted the possible models to Bayesian network structures sharing the property that all the existing arcs start from the root node representing the class variable.

Consequently, in this setup the number of possible models equals to the number of different variable subsets, and the models can be regarded as "pruned" versions of the standard Naive Bayes model where the class variable is connected to all the other variables. This allowed us to compute each model selection score exhaustively to all the possible models. Nevertheless, although we were capable of distributing the test runs to a network of Linux workstations, for pragmatic reasons we had to restrict the number of variables to be under 15; this left us with 18 UCI datasets that were used in the experiments.

As we wanted to average the results over all the 18 datasets, which were of different size, a random sample of size 500 of each dataset was used in this set of experiments. Datasets with continuous variables were discretized by using a discretization scheme based on the K-means clustering algorithm, and missing data items were replaced by a special character treated as a value of the corresponding discrete variable.

Table 1: The model selection criteria used in the experiments.

| LABEL | CRITERION |
|---|---|
| fCV | 10-fold crossvalidation as described in Section 2.2.3. |
| fCV* | 10-fold crossvalidation averaged over 10 random data orderings. |
| looCV | Leave-one-out crossvalidation as described in Section 2.2.3. |
| uEVI | The unsupervised marginal likelihood (1). |
| ~sEVI | The approximative supervised marginal likelihood (4). |
| PREQ | The supervised prequential criterion (7). |
| PREQ* | The supervised prequential criterion averaged over 10 random data orderings. |
| TRloss | The loss $L(\mathcal{D} \mid M)$ computed by evaluating the model $M$ with the training data $\mathcal{D}$ by using either the 0/1-loss or the log-loss. |

A single experiment consisted of the following steps.

1. The data ordering was first randomized while at the same time maintaining the class distributions as homogeneous as possible throughout the whole dataset, i.e., a stratified sample was created.

2. The dataset was then split into two parts (of equal size): the training data $\mathcal{D}$ and the test data $\mathcal{D}'$.

3. After this, each possible model $M$ was evaluated by using the selected scoring functions $S$, and for each $S$, the model $M(S)$ maximizing the score $S(M \mid \mathcal{D})$ was selected.

4. The actual predictive performance of a model selection method $S$ was then measured by computing the prediction loss $L(\mathcal{D}' \mid M(S))$ in the test set by using the selected model $M(S)$.

Steps 1–4 were repeated 50 times with different stratified random data orderings, and the average prediction loss was reported as the performance score of the model selection scoring function $S$ with respect to the dataset used. This whole procedure was then performed with all the 18 datasets, by using the the model scoring functions listed in Table 1.

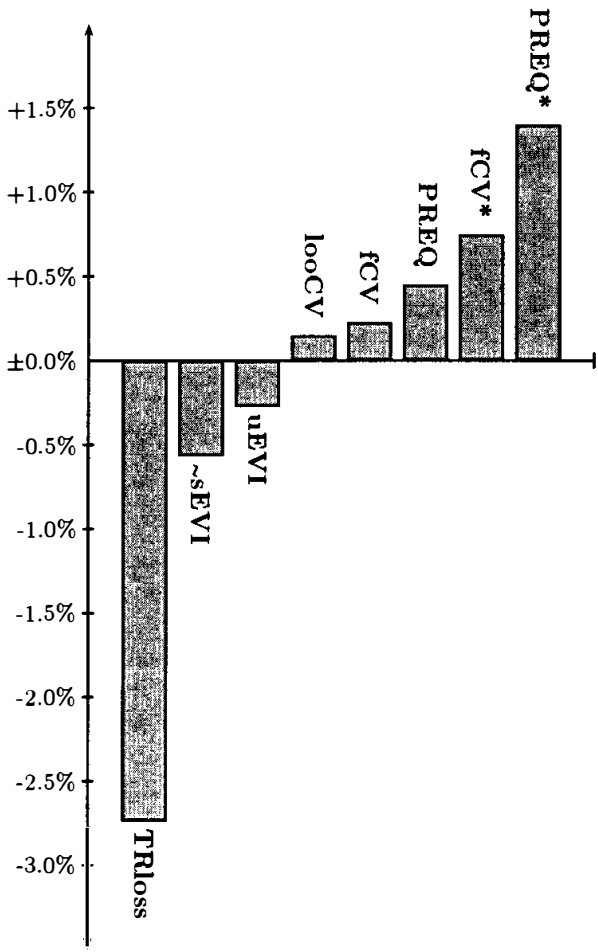

Figure 1: Average relative prediction gains with the 0/1 loss.

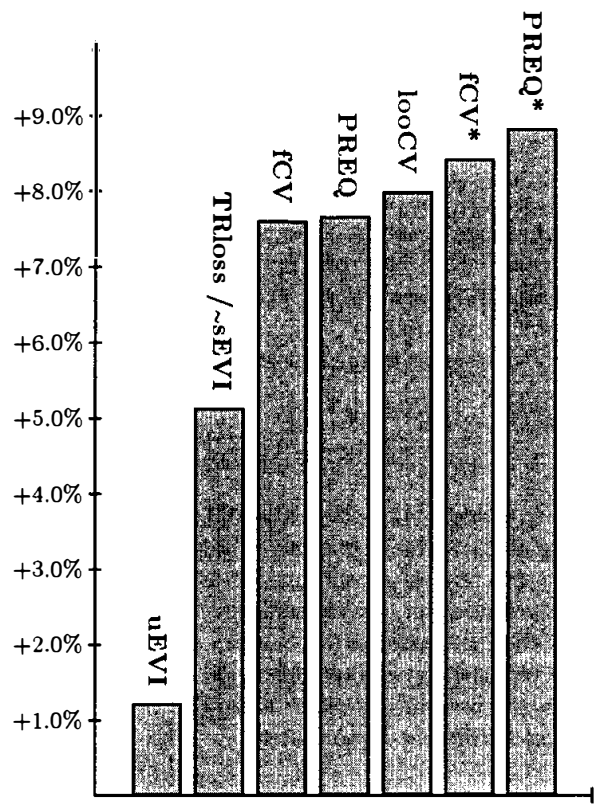

Figure 2: Average relative prediction gains with the logarithmic loss.

## 3.2 THE RESULTS

In order to be able to compare the overall performance of different methods in the 18 datasets used, the results were scaled with respect to the results obtained with the standard Naive Bayes model. In the sequel, we call the resulting score the *relative prediction gain*. This means that the relative prediction gain of the standard Naive Bayes model was always 0.0, and a relative prediction gain of, say, +1.0 means that by using the corresponding model selection method, the predictive accuracy is on the average (averaged over the 18 datasets, and over the 50 data permutations within each dataset) 1% better (the average loss is 1.0% less) than what is obtained with the standard Naive Bayes model. The average relative prediction gains computed with respect to the 0/1-loss score and the log-loss score, respectively, are shown in Figures 1 and 2. The optimal relative prediction gains (obtained by "cheating" and selecting the model after seeing the test data) in these two cases were 7.55% with the 0/1-loss, and 22.37% with the logarithmic loss.

From these results, we can make the following observations. First of all, as expected, the unsupervised marginal likelihood score turned out to perform poorly in supervised model selection tasks. As a matter fact, with the 0/1-loss the average relative prediction gain of uEVI was below zero, which means that in this case better results were obtained by using no model selection at all than by using model selection with the uEVI score. The approximative supervised marginal likelihood ~sEVI did not improve the results. We believe this is caused by the crude approximation (4), which collapses the supervised marginal likelihood into a single supervised likelihood corresponding to the TRloss method in the log-loss case.

The best results were obtained with the PREQ* method, both with the 0/1-loss and the logarithmic loss. The empirical crossvalidation schemes performed also well, especially with the log-score, which can be partly explained by the connection between crossvalidation and supervised marginal likelihood, as discussed in Section 2.2.3. It should also be noted that of the data order-dependent methods, both fCV and PREQ results improved when the corresponding score was computed by averaging over several data orderings. This is particularly noteworthy considering the success of the PREQ* method. Consequently, studying more elaborate schemes for improving the performance of the prequential approach in small sample size domains seems to offer a promising research area.

In addition to the results shown above, we also run extensive empirical tests using several other supervised and unsupervised model selection criteria, but the results were in these cases so poor that they have been excluded from the pictures in order not to distort the scaling of the histograms. However, we would like to mention that our empirical results support the observations reported in (Friedman et al., 1997), where it was discovered that the so called MDL score did not perform very well. In our setup, the relative prediction gains obtained by this criterion were $-11.38\%$ with the 0/1-loss, and $-61.11\%$ with the logarithmic loss. Nevertheless, this result cannot be considered surprising: what (Friedman et al., 1997) call the MDL score is actually only an approximation of the unsupervised marginal likelihood, also known as the *Bayesian information criterion (BIC)* or the *Schwarz criterion* (Schwarz, 1978). What is more, from the information theoretic point of view, the two-part MDL score used in (Friedman et al., 1997) can be regarded as a crude approximation of the *stochastic complexity* measure (Rissanen, 1989), for which much more elaborate formulations can be found in (Rissanen, 1996; Barron, Rissanen, & Yu, 1998). However, as discussed in (Kontkanen, Myllymäki, Silander, Tirri, & Grünwald, 1999; Friedman et al., 1997), these criteria are inherently unsupervised, and establishing supervised variants of them seems to be a difficult task.

## 4   CONCLUSIONS

We have empirically compared the performance of unsupervised and supervised model selection criteria in several supervised classification domains. The extensive empirical tests show that the supervised criteria clearly outperform the unsupervised methods, and that using a supervised model selection scheme generally improves the prediction accuracy, and is hence a reasonable thing to do in practice.

The best results were obtained by using Dawid's prequential approach, but also the empirical crossvalidation techniques showed good performance. The success of crossvalidation can be partly explained by the discussed connection between crossvalidation and the supervised marginal likelihood criterion: crossvalidation can be regarded as an approximation of this theoretical criterion which cannot be computed exactly in feasible time.

One of the most interesting empirical observations was the fact that the performance of the order-dependent methods could be improved by a very simple averaging scheme. Consequently, although asymptotically the data order bears no relevance in prequential analysis, our results suggest that development of more elaborate prequential schemes for small sample cases constitutes a practically significant research area.

The model family used in this set of experiments consisted of "pruned Naive Bayes" network structures. By restricting ourselves to this limited subset of Bayesian networks we were capable of computing the model selection criteria for all the models in the model family, thus preventing any possible bias caused by a search algorithm going through the space of all Bayesian networks. Of course, we have to admit the possibility that by focusing on this limited subfamily of Bayesian networks we may have caused a bias in the results obtained. On the other hand, it can also be argued that any search algorithm causes an implicit bias in the results, which makes a fair comparison between different model selection criteria quite difficult.

Finally, we would like to point out that the empirical setup used was designed solely for comparing different criteria in the task of choosing between alternative models — the goal was not to try to estimate the future predictive accuracy per se, which is a much more difficult task. Studying this problem was left as a topic for future work.

### Acknowledgements

This research has been supported by the Technology Development Center (TEKES), and the Academy of Finland.